\documentclass{article}
\usepackage{spconf,amsmath,amssymb,graphicx}

\DeclareMathOperator*{\argmin}{arg\!min}
\title{FACE AGING WITH CONDITIONAL GENERATIVE ADVERSARIAL NETWORKS}

\name{Grigory Antipov$^{\star \dagger}$ \qquad Moez Baccouche$^{\star}$ \qquad Jean-Luc Dugelay$^{\dagger}$}

\address{$^{\star}$ Orange Labs, 4 rue Clos Courtel, 35512 Cesson-S\'evign\'e, France \\
    $^{\dagger}$ Eurecom, 450 route des Chappes, 06410 Biot, France}

\begin{document}

\maketitle

\begin{abstract}

It has been recently shown that Generative Adversarial Networks (GANs) can produce synthetic images of exceptional visual fidelity.
In this work, we propose the GAN-based method for automatic face aging.
Contrary to previous works employing GANs for altering of facial attributes, we make a particular emphasize on preserving the original person's identity in the aged version of his/her face.
To this end, we introduce a novel approach for ``Identity-Preserving'' optimization of GAN's latent vectors.
The objective evaluation of the resulting aged and rejuvenated face images by the state-of-the-art face recognition and age estimation solutions demonstrate the high potential of the proposed method. 

\end{abstract}


\section{INTRODUCTION}
\label{sec:introduction}

Face aging, also known as age synthesis~\cite{fu2010age} and age progression~\cite{shu2015personalized}, is defined as aesthetically rendering a face image with natural aging and rejuvenating effects on the
individual face~\cite{fu2010age}.
It has plenty of applications in various domains including cross-age face recognition~\cite{park2010age}, finding lost children and entertainment~\cite{wang2016recurrent}.
Traditional face aging approaches can be roughly split between the prototyping ones~\cite{tiddeman2001prototyping,kemelmacher2014illumination} and the modeling ones~\cite{suo2010compositional,tazoe2012facial}.
Prototyping approaches estimate average faces within predefined age groups.
The differences between these faces constitute the aging patterns which are further used to transform an input face image into the target age group.
The prototype approaches are simple and fast, but since they are based on general rules, they totally discard the personalized information which results in unrealistic images.
On the contrary, modeling approaches employ parametric models to simulate the aging mechanisms of muscles, skin and skull of a particular individual.
However, these approaches often require face aging sequences of the same person with wide range of ages which are very costly to collect.

The presented traditional face aging approaches are limited to modeling of the aging patterns missing the global comprehension of a human face (its personality traits, facial expression, possible facial accessories etc.)
However, in many real-life use cases, face aging must be combined with other face alterations, such as adding sunglasses or beard.
These non-trivial modifications require global generative models of human faces.
Natural image generation has been thoroughly studied for years, but it has not been until 2014 when advances in deep learning has allowed to produce image samples and interpolations of very high visual fidelity.
The key model which has made it possible is called Generative Adversarial Network (GAN)~\cite{goodfellow2014generative} and is presented in Subsection~\ref{subsec:age_cgan}.
Contrary to autoencoders (for example, variational autoencoders~\cite{kingma2014auto}) which optimize the $L_{2}$ reconstruction loss and hence produce blurry images, GANs are explicitly trained to generate the most plausible and realistic images which are hard to distinguish from real data.
GANs have already been employed to perform modifications on human faces, such as changing the hair color, adding sunglasses or even binary aging (i.e. simply making face look older or younger without precising particular age categories)~\cite{larsen2016autoencoding,perarnau2016invertible}.
A common problem of previous GAN-based methods of face modification~\cite{larsen2016autoencoding,perarnau2016invertible} is the fact that the original person's identity is often lost in the modified image.
Therefore, in the present study, we focus on \textit{identity-preserving} face aging.
In particular, our contributions are as following:
\begin{enumerate}
	\item We design Age-cGAN (Age Conditional Generative Adversarial Network), the first GAN to generate high quality synthetic images within required age categories.
	\item We propose a novel latent vector optimization approach which allows Age-cGAN to reconstruct an input face image preserving the original person's identity.
\end{enumerate}

The rest of the paper is composed of Section~\ref{sec:proposed_approach} where we present our face aging method, Section~\ref{sec:experiments} where we evaluate the proposed identity-preserving face reconstruction and age synthesis approaches, and Section~\ref{sec:conclusions_future_works} which concludes this study and gives the directions for the future work.

\section{Proposed Method}
\label{sec:proposed_approach}

\begin{figure*}[t!]
	\begin{center}
			\includegraphics[width=\linewidth]{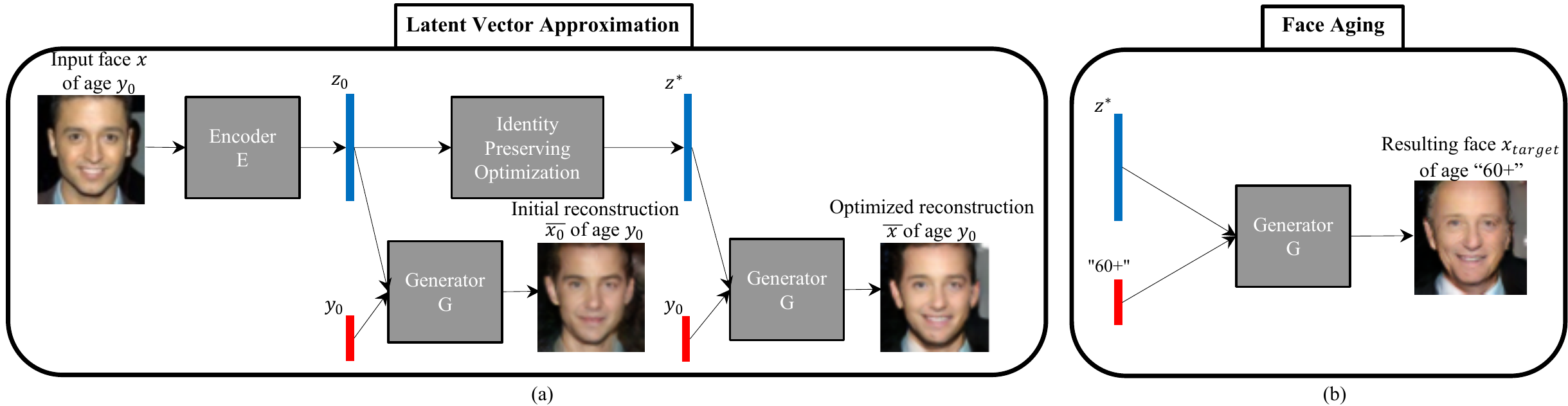}
	\end{center}
	\caption{Our face aging method. (a) approximation of the latent vector to reconstruct the input image; (b) switching the age condition at the input of the generator $G$ to perform face aging.}
	\label{fig:method_overview}
\end{figure*}

Our face aging method is based on Age-cGAN, a generative model for synthesis of human faces within required age categories.
The design of Age-cGAN is detailed in Subsection~\ref{subsec:age_cgan}.
Once Age-cGAN is trained, the face aging is done in two steps (cf. Figure~\ref{fig:method_overview}):
\begin{enumerate}
	\item Given an input face image $x$ of age $y_{0}$, find an optimal latent vector $z^{*}$ which allows to generate a reconstructed face $\bar{x}=G(z^{*},y_{0})$ as close as possible to the initial one (cf. Figure~\ref{fig:method_overview}-(a)).
	\item Given the target age $y_{target}$, generate the resulting face image $x_{target}=G(z^{*},y_{target})$ by simply switching the age at the input of the generator (cf. Figure~\ref{fig:method_overview}-(b)).
\end{enumerate}
The first step of the presented face aging method (i.e. input face reconstruction) is the key one.
Therefore, in Subsection~\ref{subsec:face_approximation_cgan}, we present our approach to approximately reconstruct an input face making a particular emphasize on preserving the original person's identity in the reconstructed image.

\subsection{Age Conditional Generative Adversarial Network}
\label{subsec:age_cgan}

Introduced in~\cite{goodfellow2014generative}, GAN is a pair of neural networks $(G,D)$: the generator $G$ and the discriminator $D$.
$G$ maps vectors $z$ from the noise space $N^{z}$ with a known distribution $p_{z}$ to the image space $N^{x}$.
The generator's goal is to model the distribution $p_{data}$ of the image space $N^{x}$ (in this work, $p_{data}$ is the distribution of all possible face images).
The discriminator's goal is to distinguish real face images coming from the image distribution $p_{data}$ and synthetic images produced by the generator.
Both networks are iteratively optimized against each other in a minimax game (hence the name ``adversarial'').

Conditional GAN (cGAN)~\cite{mirza2014conditional,gauthier2014conditional} extends the GAN model allowing the generation of images with certain attributes (``conditions'').
In practice, conditions $y\in N^{y}$ can be any information related to the target face image: level of illumination, facial pose or facial attribute.
More formally, cGAN training can be expressed as an optimization of the function $v(\theta_{G},\theta_{D})$, where $\theta_{G}$ and $\theta_{D}$ are parameters of $G$ and $D$, respectively:

\begin{equation}
	\begin{split}
		\min_{\theta_{G}}\max_{\theta_{D}}{}&v(\theta_{G},\theta_{D})=\mathbf{E}_{x,y\sim p_{data}}[\log{D(x,y)}] \\ &+\mathbf{E}_{z\sim p_{z}(z),\widetilde{y}\sim p_{y}}[\log{(1-D(G(z,\widetilde{y}),\widetilde{y}))}]
	\end{split}
	\label{eqn:cgan}
\end{equation}

The Age-cGAN model proposed in this work uses the same design for the generator $G$ and the discriminator $D$ as in~\cite{radford2016unsupervised}.
Following~\cite{perarnau2016invertible}, we inject the conditional information at the input of $G$ and at the first convolutional layer of $D$.
Age-cGAN is optimized using the ADAM algorithm~\cite{kingma2014adam} during $100$ epochs.
In order to encode person's age, we have defined six age categories: $0$-$18$, $19$-$29$, $30$-$39$, $40$-$49$, $50$-$59$ and $60+$ years old.
They have been selected so that the training dataset (cf. Subsection ~\ref{subsec:datasets}) contains at least $5,000$ examples in each age category.
Thus, the conditions of Age-cGAN are six-dimensional one-hot vectors.

\subsection{Approximative Face Reconstruction with Age-cGAN}
\label{subsec:face_approximation_cgan}

\subsubsection{Initial Latent Vector Approximation}

Contrary to autoencoders, cGANs do not have an explicit mechanism for inverse mapping of an input image $x$ with attributes $y$ to a latent vector $z$ which is necessary for image reconstruction: $x=G(z,y)$.
As in~\cite{perarnau2016invertible,zhu2016generative}, we circumvent this problem by training an encoder $E$, a neural network which approximates the inverse mapping.

In order to train $E$, we generate a synthetic dataset of $100$K pairs $(x_{i},G(z_{i},y_{i}))$, $i=1,\dots,10^{5}$, where $z_{i}\sim N(0,I)$ are random latent vectors, $y_{i}\sim U$ are random age conditions uniformly distributed between six age categories, $G(z,y)$ is the generator of the priorly trained Age-cGAN, and $x_{i}=G(z_{i},y_{i})$ are the synthetic face images.
$E$ is trained to minimize the Euclidean distances between estimated latent vectors $E(x_{i})$ and the ground truth latent vectors $z_{i}$.

Despite GANs are arguably the most powerful generative models today, they cannot \textit{exactly} reproduce the details of all real-life face images with their infinite possibilities of minor facial details, accessories, backgrounds etc.
In general, a natural input face image can be rather \textit{approximated} than \textit{exactly reconstructed} with Age-cGAN.
Thus, $E$ produce \textit{initial latent approximations} $z_{0}$ which are good enough to serve as initializations of our optimization algorithm explained hereafter.

\subsubsection{Latent Vector Optimization}

Face aging task, which is the ultimate goal of this work, assumes that while the age of a person must be changed, his/her identity should remain intact.
In Subsection~\ref{subsec:identity_preserving_face_reconstruction_aging}, it is shown that though initial latent approximations $z_{0}$ produced by $E$ result in visually plausible face reconstructions, the identity of the original person is lost in about $50$\% of cases (cf. Table~\ref{tab:face_reconstruction_scores}).
Therefore, initial latent approximations $z_{0}$ must be improved.

In~\cite{zhu2016generative}, the similar problem of image reconstruction enhancement is solved 
by optimizing the latent vector $z$ to minimize the pixelwise Euclidean distance between the ground truth image $x$ and the reconstructed image $\bar{x}$.
However, in the context of face reconstruction, the described ``Pixelwise'' latent vector optimization has two clear downsides: firstly, it increases the blurriness of reconstructions and secondly (and more importantly), it focuses on unnecessary details of input face images which have a strong impact on pixel level, but have nothing to do with a person's identity (like background, sunglasses, hairstyle, moustache etc.)

Therefore, in this paper, we propose a novel ``Identity-Preserving'' latent vector optimization approach.
The key idea is simple: given a face recognition neural network $FR$ able to recognize a person's identity in an input face image $x$, the difference between the identities in the original and reconstructed images $x$ and $\bar{x}$ can be expressed as the Euclidean distance between the corresponding embeddings $FR(x)$ and $FR(\bar{x})$.
Hence, minimizing this distance should improve the identity preservation in the reconstructed image $\bar{x}$:

\begin{equation}
	{z^{*}}_{IP}=\argmin_{z}{||FR(x)-FR(\bar{x})||_{L_{2}}}
	\label{eqn:FR_optim}
\end{equation}

In this paper, $FR$ is an internal implementation of the ``FaceNet'' CNN~\cite{schroff2015facenet}.
The generator $G(z,y)$ and the face recognition network $FR(x)$ are differentiable with respect to their inputs, so the optimization problem~\ref{eqn:FR_optim} can be solved using the L-BFGS-B algorithm~\cite{byrd1995limited} with backtracking line search.
The L-BFGS-B algorithm is initialized with initial latent approximations $z_{0}$.
Here and below in this work, we refer to the results of ``Pixelwise'' and ``Identity-Preserving'' latent vector optimizations as \textit{optimized latent approximations} and denote them respectively as ${z^{*}}_{pixel}$ and ${z^{*}}_{IP}$.
In Subsection~\ref{subsec:identity_preserving_face_reconstruction_aging}, it is shown both subjectively and objectively that ${z^{*}}_{IP}$ better preserves a person's identity than ${z^{*}}_{pixel}$.

\section{EXPERIMENTS}
\label{sec:experiments}

\subsection{Dataset}
\label{subsec:datasets}

Age-cGAN has been trained on the \textit{IMDB-Wiki\_cleaned} dataset~\cite{antipov2016apparent} of about $120$K images which is a subset of the public \textit{IMDB-Wiki} dataset~\cite{rothe2015dex}.
More precisely, $110$K images have been used for training of Age-cGAN and the remaining $10$K have been used for the evaluation of identity-preserving face reconstruction (cf. Subsection~\ref{subsec:identity_preserving_face_reconstruction_aging}).

\subsection{Age-Conditioned Face Generation}
\label{subsec:age_conditioned_face_generation}

\begin{figure}[h!]
	\begin{center}
			\includegraphics[width=\linewidth]{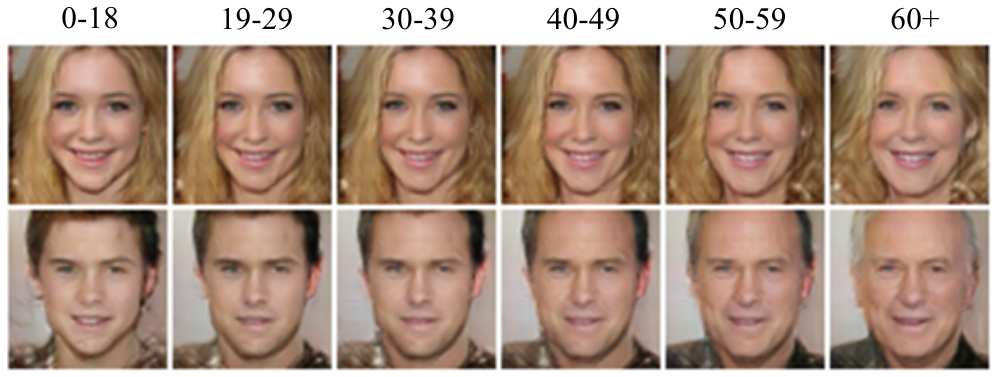}
	\end{center}
	\caption{Examples of synthetic images generated by our Age-cGAN using two random latent vectors $z$ (rows) and conditioned on the respective age categories $y$ (columns).}
	\label{fig:face_synthesis}
\end{figure}

Figure~\ref{fig:face_synthesis} illustrates synthetic faces of different ages generated with our Age-cGAN.
Each row corresponds to a random latent vector $z$ and six columns correspond to six age conditions $y$.
Age-cGAN perfectly disentangles image information encoded by latent vectors $z$ and by conditions $y$ making them independent.
More precisely, we observe that latent vectors $z$ encode person's identity, facial pose, hair style, etc., while $y$ encodes \textit{uniquely} the age.

\begin{figure*}[t!]
	\begin{center}
			\includegraphics[width=\linewidth]{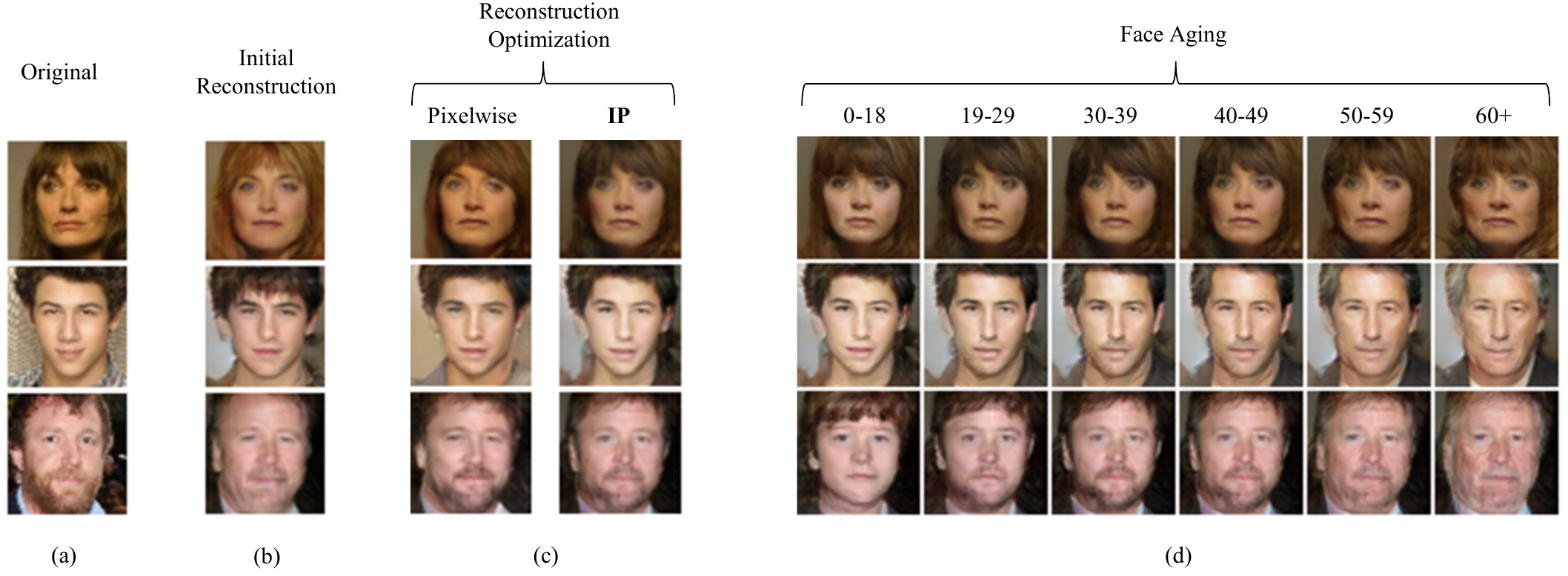}
	\end{center}
	\caption{Examples of face reconstruction and aging. (a) original test images, (b) reconstructed images generated using the initial latent approximations: $z_{0}$, (c) reconstructed images generated using the ``Pixelwise'' and ``Identity-Preserving'' optimized latent approximations: ${z^{*}}_{pixel}$ and ${z^{*}}_{IP}$, and (d) aging of the reconstructed images generated using the identity-preserving ${z^{*}}_{IP}$ latent approximations and conditioned on the respective age categories $y$ (one per column).}
	\label{fig:face_reconstruction}
\end{figure*}

In order to objectively measure how well Age-cGAN manages to generate faces belonging to precise age categories, we have used the state-of-the-art age estimation CNN described in~\cite{antipov2016apparent}.
We compare the performances of the age estimation CNN on real images from the test part of \textit{IMDB-Wiki\_cleaned} and on $10$K synthetic images generated by Age-cGAN.
Despite the age estimation CNN has never seen synthetic images during the training, the resulting mean age estimation accuracy on synthetic images is just $17$\% lower than on natural ones.
It proves that our model can be used for generation of realistic face images with the required age.

\subsection{Identity-Preserving Face Reconstruction and Aging}
\label{subsec:identity_preserving_face_reconstruction_aging}

As explained in Subsection~\ref{subsec:face_approximation_cgan}, we perform face reconstruction (i.e. the first step of our face aging method) in two iterations: firstly, (1) using initial latent approximations obtained from the encoder $E$ and then (2) using optimized latent approximations obtained by either ``Pixelwise'' or ``Identity-Preserving'' optimization approaches.
Some examples of original test images, their initial and optimized reconstructions are presented in Figure~\ref{fig:face_reconstruction} ((a), (b) and (c), respectively).

It can be seen in Figure~\ref{fig:face_reconstruction} that the optimized reconstructions are closer to the original images than the initial ones.
However, the choice is more complicated when it comes to the comparison of the two latent vector optimization approaches.
On the one hand, ``Pixelwise'' optimization better reflects superficial face details: such as the hair color in the first line and the beard in the last line.
On the other hand, the identity traits (like the form of the head in the second line or the form of the eyes in the first and last lines) are better represented by ``Identity-Preserving'' optimization.

For the sake of objective comparison of the two approaches for the identity-preserving face reconstruction, we employ ``OpenFace''~\cite{amos2016openface} software which is currently one of the best open-source face recognition solutions.
Given two face images, ``OpenFace'' decides whether they belong to the same person or not.
Using both approaches, we have reconstructed $10$K test images of the \textit{IMDB-Wiki\_cleaned} dataset and fed the resulting images alongside corresponding originals to ``OpenFace''.
Table~\ref{tab:face_reconstruction_scores} presents the percentages of ``OpenFace'' positive outputs (i.e. when the software believed that a face image and its reconstruction belong to the same person).
The results confirm the visual observations presented above.
Initial reconstructions allow ``OpenFace'' to recognize the original person only in half of test examples.
This percentage is slightly increased by ``Pixelwise'' optimization but the improvement is marginal.
On the contrary, ``Identity-Preserving'' optimization approach preserves the person's identities much better demonstrating by far the best face recognition performance of $82.9$\%.

\begin{table}
	\centering
	\small
		\begin{tabular}{|c|c|}
		\hline
		Reconstruction type & FR score\\
		\hline
		\hline
		Initial Reconstruction (${z_{0}}$) & $53.2$\% \\
		\hline
		\hline
		``Pixelwise'' Optimization (${z^{*}}_{pixel}$) & $59.8$\% \\
		\hline
		``Identity-Preserving'' Optimization (${z^{*}}_{IP}$) & \textbf{82.9}\% \\
		\hline
		\end{tabular}
	\caption{``OpenFace'' Face Recognition (FR) scores on three compared types of face reconstruction.}
	\label{tab:face_reconstruction_scores}
\end{table}

Finally, once an identity-preserving face reconstruction $\bar{x}=G({z^{*}}_{IP},y_{0})$ of the original image $x$ is obtained, we can simply substitute the initial age category $y_{0}$ by the target age category $y_{target}$ in order to obtain the output of our face aging method: $x_{target}=G({z^{*}}_{IP},y_{target})$.
Figure~\ref{fig:face_reconstruction}-(d) illustrates how it works in practice.
Our method manages both to realistically age an originally young face into a senior one as in the 2nd line of Figure~\ref{fig:face_reconstruction} and vice-versa as in the last line.

\section{CONCLUSIONS AND FUTURE WORK}
\label{sec:conclusions_future_works}

In this work, we have proposed a new effective method for synthetic aging of human faces based on Age Conditional Generative Adversarial Network (Age-cGAN).
The method is composed of two steps: (1) input face reconstruction requiring the solution of an optimization problem in order to find an optimal latent approximation $z^{*}$, (2) and face aging itself performed by a simple change of condition $y$ at the input of the generator.
The cornerstone of our method is the novel ``Identity-Preserving'' latent vector optimization approach allowing to preserve the original person's identity in the reconstruction.
This approach is universal meaning that it can be used to preserve identity not only for face aging but also for other face alterations (e.g. adding a beard, sunglasses etc.)

Our face aging method can be used for synthetic augmentation of face datasets and for improving the robustness of face recognition solutions in cross-age scenarios.
It is part of our future work.
Moreover, we believe that the face reconstruction part of our method can be further improved by combining ``Pixelwise'' and ``Identity-Preserving'' approaches into one optimization objective.
We are also planning to explore this path in our further studies.

\bibliographystyle{IEEEbib}
\bibliography{icip_2017_manuscript_refs}

\end{document}